\documentclass[conference,a4paper]{APSIPA2026}
\usepackage[T1]{fontenc}
\usepackage{amsmath}
\usepackage{amssymb}
\usepackage{graphicx}
\usepackage{subcaption}
\usepackage{multirow}
\usepackage{threeparttable}
\usepackage{booktabs}
\usepackage{color}
\usepackage{tikz}
\usetikzlibrary{positioning,arrows.meta,fit,calc,shapes.geometric}
\usepackage{url}
\usepackage[backend=biber,style=ieee,]{biblatex}
\usepackage{geometry}
\usepackage{fancyhdr}

\fancypagestyle{firststyle}{
  \fancyhf{}
  \fancyhead[C]{2026 Asia Pacific Signal and Information Processing Association Annual Summit and Conference (APSIPA ASC)}
}

\begin{document}

\title{Style-Driven Data Synthesis and Degradation-Aware Enhancement for Ultrasound Image Restoration}

\author{
\authorblockN{
Yu-Kai Wang\authorrefmark{1},
Chun-Xin Tan\authorrefmark{1},
Manh-Hung Nguyen\authorrefmark{2}, and
Ching-Chun Huang\authorrefmark{1}
}

\authorblockA{
\authorrefmark{1}
Department of Computer Science, National Yang Ming Chiao Tung University, Hsinchu, Taiwan \\
E-mail: \{jason0411202.cs14, tanchunxin2001.cs14, chingchun\}@nycu.edu.tw}

\authorblockA{
\authorrefmark{2}
Ho Chi Minh City University of Technology and Engineering, Ho Chi Minh City, Vietnam \\
E-mail: hungnm@hcmute.edu.vn}
}

\maketitle
\thispagestyle{firststyle}
\pagestyle{fancy}
\fancyhf{}

\begin{abstract}
Low-cost handheld ultrasound devices can be widely deployed compared to professional hospital ultrasound machines. However, their images suffer from compound degradation that can mislead clinical judgment. Motivated by this observation, mapping handheld low-quality (LQ) to hospital high-quality (HQ) images has been considered a valuable research question. Conventionally, the mapping requires pixel-aligned LQ–HQ pairs. This requirement is unsatisfactory in practical scenarios because real scans at different times are never pixel-aligned. This paper addresses the challenge with a two-stage framework. The first stage generates pixel-aligned LQ--HQ datasets, and the second stage trains an enhancement model that improves LQ images. The first stage trains a cycle-consistent style-transfer model on unaligned real LQ-HQ pairs to learn a HQ$\rightarrow$LQ model. Then, the model transforms real HQ images into pixel-aligned LQ images. Based on the dataset generated by the first stage, the second stage uses the Dual Degradation-Guided (DDG) Low-Rank Adaptation (LoRA) method to fine-tune an LQ$\rightarrow$HQ model based on aligned pairs. In this stage, the model is based on the well known PiSA-SR framework but inserts a degradation-conditioned correction matrix. 
% Because there are various degradations among handheld ultrasound devices, organs, and patients, a single fixed update in conventional LoRA is insufficient. 
Experimental results on the USenhance2023 dataset show that the FID metric is improved by 16.7\% over the strongest baseline while other metrics indicate that our enhanced outputs are well aligned with the real HQ distribution. The source code of our method is available at \url{https://github.com/Jason0411202/DDG_LoRA}.\end{abstract}

%================================================================
\section{Introduction}
%================================================================

Handheld ultrasound devices have become the default imaging tool in rural hospitals and primary-care clinics where professional ultrasound machines are unavailable or too expensive~\cite{zhou2020_handheld}. Their images, however, are degraded by strong speckle noise, motion-like blur, and compressed contrast. In the natural-image domain, deep learning models~\cite{wang2018_esrgan} have demonstrated considerable success in generating {high-quality (HQ) images from low-quality (LQ) inputs}. This raises a natural question: could the advanced techniques improve handheld ultrasound quality enough to reduce the urgency of costly hardware upgrades?

Supervised LQ$\rightarrow$HQ enhancement requires pixel-aligned LQ--HQ pairs to train the enhancement model. However, scanning the same patient with both device types inevitably introduces differences in probe-angle and tissue-deformation. The LQ--HQ pairs are never pixel-aligned. Training an enhancement model with unaligned data makes the supervision signal inconsistent and leads to poor convergence. One natural approach is to synthesize aligned pairs by applying hand-crafted degradation to HQ images~\cite{realesrgan}. This approach ensures pixel-level alignment, but the degradation recipes are too idealized to reproduce what real handheld devices exhibit in practice. Another approach is physics-based simulation of the acoustic imaging process~\cite{fieldii}. However, the dynamic and device-dependent factors make accurate simulation prohibitively complex. 

An alternative solution is to directly learn LQ$\rightarrow$HQ enhancers from an unpaired dataset using a CycleGAN-style framework~\cite{cyclegan}. However, the generator must produce the HQ side without any real HQ supervision, which often results in unacceptable failures on medical images.
\begin{figure}[t]
\centering
\includegraphics[width=\columnwidth]{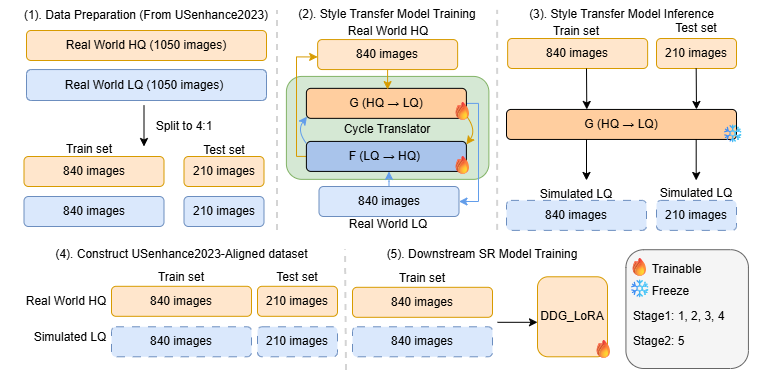}
\caption{Style-driven data synthesis pipeline. Stage~1 splits USenhance2023 into a 4:1 train/test partition (1), trains a style transfer model $G$ on unaligned real LQ--HQ pairs (2), applies $G$ to each real HQ image $y_i$ to obtain a pixel-aligned LQ counterpart $\tilde{x}_i$ (3), and assembles the USenhance2023-Aligned dataset (4). Stage~2 trains the DDG-LoRA enhancer on this dataset (5).}
\label{fig:pipeline}
\end{figure}

This paper proposes a two-stage framework that handles the alignment and domain-gap problems separately (Fig.~\ref{fig:pipeline}). Stage one learns HQ$\rightarrow$LQ from real unaligned LQ--HQ pairs to synthesize aligned pairs, and stage two then trains an LQ$\rightarrow$HQ enhancement model on those pairs. In the first stage, we use a style-transfer CycleDiff~\cite{cyclediff} as the degrader. Unlike conventional CycleGAN that translates images through a single generator and tends to overfit individual data points rather than covering the full distribution of both domains, frequently causing structural distortion, CycleDiff integrates the translation process into a diffusion framework at a deeper level. A denoising network predicts the image component at each time step and helps to isolate the clean signal from the noisy input. Therefore, the synthesized LQ is pixel-aligned with the source HQ. Moreover, the HQ images themselves are never modified, so the supervision target remains real clinical data. This eliminates the hallucination risk inherent in methods that generate the HQ side directly.

In the second stage, a LQ$\rightarrow$HQ enhancer is trained from the data generated in stage one. The enhancer is expected not only to address denoising or sharpening, but also to improve speckle statistics and tissue-texture changes that are critical in ultrasound analysis. The requirements are far from what CNN-based SR models~\cite{edsr} can express. Therefore,  this paper uses PiSA-SR~\cite{pisasr}, which is a powerful framework, as the backbone for our enhancer. Its dual-LoRA design splits the conflicting objectives of low-level pixel fidelity and high-level semantic plausibility into two branches. However, PiSA-SR focuses on super resolution but is not designed to model complex degradation as seen in ultrasound images. Recently, a degradation-conditioned correction matrix was introduced in S3Diff~\cite{s3diff} to model degradation effectively. Therefore, we integrate the concept of a degradation-conditioned correction matrix into PiSA-SR to train the enhancer. The combination of the degradation guide in S3Diff and the dual-LoRA PiSA-SR creates a Dual Degradation-Guided (DDG) LoRA method that enhances LQ images with complex speckle noise.

In summary, the contributions of the paper are as follows:

\begin{itemize}
\item We propose a two-stage pipeline to train a robust enhancer from unpaired data. The first stage is a cycle-consistent style-transfer model to learn HQ$\rightarrow$ LQ from unaligned real clinical pairs. It provides pixel-aligned synthetic LQ--HQ data to train the enhancer in the second stage. The HQ images are never modified and ensure the supervision target is free of anatomical hallucination.
\item  We extend PiSA-SR with a degradation-conditioned correction matrix $C(\boldsymbol{d})$ inserted between
$A$ and $B$ in both LoRA branches. This input-adaptive update enables the model to handle the complex, spatially varying speckle noise characteristic of handheld ultrasound.
\item  We report diverse no-reference metrics on the real-world USenhance2023 dataset. We focus particularly on the FID metric. It measures the distributional distance between enhanced outputs and authentic HQ scans. \textcolor{black}{Our method achieves a 16.7\% better FID than the strongest baseline.} The result confirms the complementary role in handling degradation on ultrasound images.
\end{itemize}

%================================================================
\section{Related Work}
%================================================================

\subsection{Image Super-Resolution and Degradation Modeling}

Deep super-resolution has progressed from CNN-based methods such as EDSR~\cite{edsr} to real-world SR pipelines, and more recently to diffusion-based models. Within the real-world SR line, each method adopts a different degradation assumption, where Real-ESRGAN~\cite{realesrgan} uses high-order degradation, DARSR~\cite{darsr} performs test-time adaptation, and PDM~\cite{pdm} adopts a probabilistic degradation model. On the diffusion side, OSEDiff~\cite{osediff} distills a latent diffusion model into a one-step restorer, and PiSA-SR introduces dual-LoRA branches for adjustable fidelity--perception trade-offs. Across these paradigms, performance on real inputs still hinges on whether the assumed degradation matches the real one; when it does not, architectural refinement alone cannot close the gap.

\subsection{Ultrasound-Specific Image Enhancement}

A similar picture emerges in ultrasound-specific enhancement. Zhou et al.~\cite{zhou2020_handheld} proposed GAN-based architectures for handheld ultrasound. Jiang et al.~\cite{usbsr} designed a two-stage wavelet-decomposition degradation pipeline. Khan et al.~\cite{khan2026_blind_us} proposed a self-supervised physics-guided degradation model. Despite their architectural differences, all of them still train on synthetically degraded LQ images, which limits how well they generalize to real handheld output.

\subsection{Unpaired Translation: Enhancement vs.\ Degradation}

CycleGAN~\cite{cyclegan}-based unpaired translation has been widely applied in ultrasound. Huang et al.~\cite{huang2022_cyclegan_us} adopted it for cross-center ultrasound domain normalization. However, an unconstrained generator used as an enhancer can create plausible-looking anatomy that is not really there. This hallucination problem is unacceptable in medical imaging.

Our work takes the opposite direction. We repurpose unpaired translation as a degradation simulator in the HQ$\rightarrow$LQ direction. The generator only needs to simulate quality loss, and the HQ training targets remain untouched real data. This removes the hallucination risk on the HQ side, while still exploiting the generator's ability to learn complex, device-specific degradation from unaligned data.

%================================================================
\section{Proposed Method}
%================================================================
Section \ref{subsec: S1} introduces how the pixel-aligned dataset is created, and Section \ref{subsec: S2} discusses how the enhancer is trained.
\subsection{Style-Driven Data Synthesis Pipeline}
\label{subsec: S1}
Given a subset $\{{x}_i\}_{i=1}^{N}$ and $\{y_i\}_{i=1}^{N}$ from the USenhance2023 dataset, the CycleDiff~\cite{cyclediff} method is used to train a style transfer model $G$. Later, $\{(\tilde{x}_i=G(y_i), y_i)\}_{i=1}^{N}$ datasets are generated to prepare for the second stage.
% CycleGAN learns the cross-domain mapping as a single forward pass through a generator,
% calling the network once per image. This one-step mapping tends to overfit individual
% data points rather than covering the full data distribution of both domains, often
% leading to mode collapse or structural distortion in the translated output. CycleDiff
% addresses this limitation through two key innovations: a decoupled diffusion formulation
% that isolates clean image components from noise, and a multi-step translation that
% operates on these components iteratively at every denoising step.

Concretely, CycleDiff models the forward diffusion in domain~$\mathcal{S}$ as
\begin{equation}
  x^{\mathcal{S}}_t = x^{\mathcal{S}}_0 + \int_0^t C^{\mathcal{S}}_t\,dt + t\,\epsilon^{\mathcal{S}},
  \quad \epsilon^{\mathcal{S}} \sim \mathcal{N}(0, I),
\end{equation}
where $C^{\mathcal{S}}_t$ is the gradient of the image attenuation process. Since the image must attenuate to zero at $t\!=\!1$, we have $C^{\mathcal{S}}_t = -x^{\mathcal{S}}_0$. A denoising network then predicts $C^{\mathcal{S}}_{t\theta}$  which represents the clean signal content isolated from the noisy mixture~$x^{\mathcal{S}}_t$. Unlike the noisy estimates produced by standard diffusion models, these image components are sufficiently clean to serve as input for the subsequent translation process. Later, the cycle consistency constraint in CycleGAN~\cite{cyclegan} is used to train style transfer.

% The translation is then performed on $C^{\mathcal{S}}_{t\theta}$ at every denoising step~$t$ via a time-dependent translation network~$G_\phi$, producing the translated component $G_\phi(C^{\mathcal{S}}_{t\theta}, t)$. The cycle consistency constraint is enforced directly on these components:
% \begin{equation}
%   C^{\mathcal{S}}_{t\theta}
%   \;\xrightarrow{G_\phi}\;
%   G_\phi(C^{\mathcal{S}}_{t\theta}, t)
%   \;\xrightarrow{F_\psi}\;
%   F_\psi\bigl(G_\phi(C^{\mathcal{S}}_{t\theta}, t),\, t\bigr)
%   \;\approx\; C^{\mathcal{S}}_{t\theta},
% \end{equation}
% so structural fidelity is preserved at every diffusion step rather than only at the final output. This iterative, step-wise translation enables CycleDiff to model the entire data distribution of both domains more faithfully than CycleGAN's one-step mapping, which is precisely why the resulting generator~$G$ produces spatially consistent output that preserves the structural layout of its input a property our pipeline relies on to guarantee pixel-level alignment between the synthesized LQ and the source HQ images.

\subsection{DDG-LoRA Enhancer}
\label{subsec: S2}

Our enhancement model is built on PiSA-SR~\cite{pisasr}, whose strong diffusion prior and dual-LoRA mechanism provide a suitable starting point for restoring the compound degradations of handheld ultrasound. PiSA-SR attaches two LoRA modules to the frozen diffusion UNet and trains them in two stages. A pixel-level LoRA $\Delta\theta_{\text{pix}}$ is trained first under an $\ell_2$ reconstruction loss for pixel-level fidelity. A semantic-level LoRA $\Delta\theta_{\text{sem}}$ is trained afterwards under a combined LPIPS~\cite{lpips} and classifier score distillation ($\mathcal{L}_{\text{CSD}}$) objective, while $\Delta\theta_{\text{pix}}$ stays frozen. Both LoRA modules follow the standard LoRA form, in which a frozen base weight $W\!\in\!\mathbb{R}^{m\times n}$ is replaced by
\begin{equation}
  W_{\text{LoRA}} = W + AB,\quad A\!\in\!\mathbb{R}^{m\times r},\ B\!\in\!\mathbb{R}^{r\times n},
  \label{eq:lora}
\end{equation}
with $r\!\ll\!\min(m,n)$, so only the two small factorizations are trainable while the UNet and VAE weights remain frozen. At inference, both LoRA modules are active and combined in a classifier-free-guidance style,
\begin{equation}
  \epsilon_\theta(z_L) = \lambda_{\text{pix}}\,\epsilon_{\theta_{\text{pix}}}(z_L)
  + \lambda_{\text{sem}}\bigl(\epsilon_{\theta_{\text{PiSA}}}(z_L) - \epsilon_{\theta_{\text{pix}}}(z_L)\bigr),
  \label{eq:fuse}
\end{equation}
where $\lambda_{\text{pix}}$ together with $\lambda_{\text{sem}}$ trade off pixel fidelity against perceptual quality.

\begin{figure}[t]
\centering
\includegraphics[width=\columnwidth]{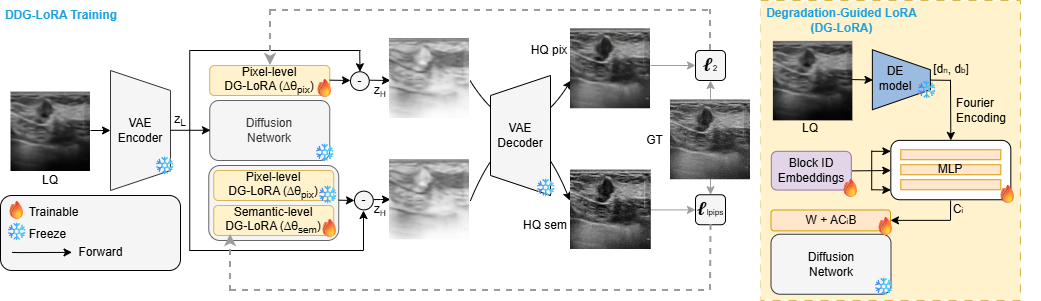}
\caption{DDG-LoRA architecture. The pixel stage trains $\Delta\theta_{\text{pix}}$, then the semantic stage freezes it and trains $\Delta\theta_{\text{sem}}$. Within each stage, DDG-LoRA replaces the static LoRA update with a degradation-conditioned variant driven by a descriptor $\boldsymbol{d}$ that the DE network extracts from the LQ image $x_L$. The semantic stage further drops the CSD loss $\mathcal{L}_{\text{CSD}}$.}
% \label{fig:arch}
\label{fig:arch}
\end{figure}

PiSA-SR was designed around natural images, and a visible cross-domain gap remains when transferred to ultrasound restoration. We introduce two targeted modifications to close this gap (Fig.~\ref{fig:arch}). The first addresses how each LoRA module responds to the input. Under Eq.~\eqref{eq:lora}, a single pair $(A, B)$ defines one fixed update $AB$ reused for every input---a poor fit for handheld ultrasound, where noise level and blur vary sharply across devices, organs, and patients. Following the S3Diff~\cite{s3diff} guidance principle, we make the update input-dependent by inserting a correction matrix $C(\boldsymbol{d})$ between $A$ and $B$. A lightweight degradation-estimation (DE) network predicts a two-dimensional descriptor $\boldsymbol{d} = [d_n, d_b]^{\!\top}\!\in\!\mathbb{R}^{2}$ from the LQ input, where $d_n$ and $d_b$ capture noise and blur levels, respectively. We encode $\boldsymbol{d}$ with Fourier features and fuse it with a layer-specific block embedding,
\begin{equation}
  \gamma(\boldsymbol{d}) = \bigl[\sin(2\pi W_e \boldsymbol{d});\ \cos(2\pi W_e \boldsymbol{d})\bigr] + \boldsymbol{e}_{\text{block}},
  \label{eq:fourier}
\end{equation}
where $W_e\!\in\!\mathbb{R}^{k\times 2}$ is a learnable projection and $\boldsymbol{e}_{\text{block}}\!\in\!\mathbb{R}^{2k}$ is a block-ID embedding that lets each UNet layer interpret the same $\boldsymbol{d}$ in a position-aware way. An MLP then maps $\gamma(\boldsymbol{d})$ to an $r\times r$ correction matrix,
\begin{equation}
  C(\boldsymbol{d}) = \operatorname{reshape}\bigl(\text{MLP}(\gamma(\boldsymbol{d}))\bigr) \in \mathbb{R}^{r\times r},
  \label{eq:cmat}
\end{equation}
which is placed between $A$ and $B$ to obtain the degradation-guided LoRA (DG-LoRA) weight update,
\begin{equation}
  W_{\text{DG-LoRA}}(\boldsymbol{d}) = W + A\,C(\boldsymbol{d})\,B.
  \label{eq:dglora}
\end{equation}
To stay compatible with PiSA-SR's dual-LoRA design, each LoRA module is replaced with its own DG-LoRA block. The Fourier projection, block embedding, and MLP are instantiated independently for each branch, yielding a pixel-level correction $C^{(p)}(\boldsymbol{d})$ and a semantic-level correction $C^{(s)}(\boldsymbol{d})$ that reshape their respective weight updates in their own way.

The second modification addresses PiSA-SR's Stage-2 objective. The text-prompt extractor behind $\mathcal{L}_{\text{CSD}}$ is trained on natural images; on ultrasound scans it produces out-of-domain concepts that carry little useful information for the diffusion prior. We therefore disable $\mathcal{L}_{\text{CSD}}$ and its text conditioning when training DDG-LoRA, reducing the Stage-2 objective to the LPIPS term alone.

\section{Experiments}
%================================================================

\subsection{Experimental Settings}

% \noindent\textbf{Dataset construction.} 
We partition the USenhance2023 collection~\cite{usenhance2023} into 840 training and 210 test samples. The collection contains 1,050 patient-matched yet unaligned clinical ultrasound pairs across five organs. Each pair couples \textcolor{black}{an LQ scan with an HQ scan, both at $256{\times}256$ resolution}. To establish pixel-aligned supervision without requiring prior spatial correspondence, we train a CycleDiff-based style-transfer degrader $G$ on the 840 unaligned pairs through three stages. Stage~1 fits a per-domain AutoencoderKL initialized from kl-f4 weights with learning rate decaying from $5{\times}10^{-6}$ to $10^{-6}$ over 50k steps \textcolor{black}{at an effective batch size of 16}, producing the per-domain latent spaces used by the later stages. Stage~2 freezes those latent spaces and trains a per-domain unconditional LDM within each via an EDM-preconditioned DhariwalUNet with learning rate decaying from $10^{-4}$ to $10^{-5}$ over 400k steps \textcolor{black}{at an effective batch size of 12}. Stage~3 stacks a timestep-conditioned ResNet generator together with an N-layer PatchGAN above the two LDMs from Stage~2 to obtain $G$, with translator learning rate $10^{-4}$ and LDM fine-tune learning rate $10^{-5}$ over 160k steps \textcolor{black}{at an effective batch size of 24}. The optimized $G$ is then applied to the 840 HQ training images to synthesize aligned LQ counterparts \textcolor{black}{at the same $256{\times}256$ resolution}, forming the USenhance2023-Aligned training set. \textcolor{black}{A 210-pair USenhance2023-Aligned test set is also produced for the held-out split.} Because $G$ only synthesizes the LQ side, the target HQ images remain unmodified throughout to keep the supervision target free of anatomical hallucination, while the synthesized LQ side faithfully reproduces device-specific degradation.

% \noindent\textbf{DDG-LoRA Training settings.} 
On the USenhance2023-Aligned set, DDG-LoRA is trained with SD-2.1-base as the pretrained backbone and warm-started by the dual LoRA checkpoints provided by PiSA-SR~\cite{pisasr}. \textcolor{black}{The synthesized LQ images are bicubically downsampled by $4{\times}$ to $64{\times}64$ to match the standard $4{\times}$ super-resolution setting shared by all baselines.} Training adopts single-step diffusion with residual learning, with a learning rate $5{\times}10^{-5}$ over 12,500 steps \textcolor{black}{at an effective batch size of 2}. The first 4,000 steps optimize only the pixel LoRA before switching to the semantic LoRA. \textcolor{black}{At inference, $\lambda_{\text{pix}}$ and $\lambda_{\text{sem}}$ are both set to the default value of $1$.}

% \noindent\textbf{Evaluation metrics.} 
\textcolor{black}{The reported experimental results are computed on the 210 unaligned real-world LQ--HQ pairs from the USenhance2023 test split, rather than directly using the pixel-aligned synthetic USenhance2023-Aligned test set.} \textcolor{black}{Because pixel-aligned ground truth (GT) is unavailable, the proposed method is evaluated with five widely used metrics that do not depend on pixel-level correspondence.} FID~\cite{fid} measures the distributional discrepancy between enhanced outputs and authentic HQ scans in a deep feature space, where lower values indicate closer alignment with real clinical data. NIQE~\cite{niqe} scores how closely an output follows natural-scene statistics, with lower values denoting a more natural appearance. PI (perceptual index)~\cite{blau2018_pirm} aggregates no-reference statistics into a single perceptual score, for which lower is better. Tenengrad~\cite{krotkov1988_focusing} reflects gradient-based sharpness and rewards images with well-preserved anatomical edges. Entropy quantifies the richness of structural details, with higher values indicating more informative content. \textcolor{black}{Among the five metrics, FID is the most important for ultrasound enhancement. Because FID is the only metric that compares the enhanced outputs against the real HQ distribution, it reveals over-sharpened yet structurally hallucinated artifacts that non-reference metrics fail to detect.}

\subsection{Impact of Training Data on Enhancement Performance}
This section aims to evaluate the effect of training data on DDG-LoRA. We construct five paired training datasets, each generated by a different degradation pipeline, and train DDG-LoRA on each dataset independently. realesrgan\_deg~\cite{realesrgan} is a widely adopted general-purpose hand-crafted pipeline originally developed for natural images and now serves as a common reference for synthesized degradations across imaging modalities. physics\_guided\_deg~\cite{khan2026_blind_us} and usbsr\_deg~\cite{usbsr} are hand-crafted pipelines proposed in prior ultrasound enhancement studies and tailored to the ultrasound domain. The former simulates ultrasound-specific artifacts from acoustic physics, while the latter operates in the frequency domain to capture characteristic ultrasound noise. ultrasound\_deg is another hand-crafted pipeline that we propose for comparison, combining blur, downsampling, Gaussian noise, and speckle noise. Finally, USenhance2023-Aligned is our style-driven training set learned directly from real handheld LQ scans.

% To evaluate whether realistic synthesized degradation translates to better downstream enhancement, we assess model performance across both controlled simulated environments and real-world clinical settings.

% We first evaluate the adaptivity of our style-driven degrader using a simulated environment where full-reference evaluation is possible. We construct a target test set by applying the frequency-domain degradation pipeline from USBSR~\cite{usbsr} to the HQ images. 

Table~\ref{tab:ddglora} presents the quantitative results in a real-world clinical setting. Among the five training sources, USenhance2023-Aligned (ours) ranks first on FID, PI, Tenengrad, and Entropy, and second on NIQE. The FID drops from 127.01, the lowest-FID hand-crafted baseline given by \textcolor{black}{realesrgan\_deg}, to 99.00, indicating that the enhanced outputs lie closer to the distribution of authentic HQ scans. \textcolor{black}{This finding suggests that learning the degradation directly from real handheld scans captures statistical characteristics that hand-crafted pipelines cannot reproduce.}

% Building upon the simulated results, we isolate the effect of the training data in a real-world clinical setting, where pixel-aligned ground truth is unavailable. We fix the DDG-LoRA backbone and vary only the degradation pipeline used to generate the training LQ images. As presented in Table~\ref{tab:ddglora}, models trained on our style-driven \texttt{USenhance2023-Aligned} dataset achieve the best scores across all five no-reference metrics. The improvement in speckle suppression is particularly evident; the ENL achieved by our dataset is more than double that of the strongest hand-crafted baseline. Visual comparisons (Fig.~\ref{fig:visual_ddglora}) corroborate this, showing cleaner tissue backgrounds and sharper anatomical boundaries.

\begin{table}[t]
\begin{center}
\caption{Comparison of DDG-LoRA (ours) trained with different LQ data sources, evaluated on the real-world USenhance2023 test set. \textcolor{red}{Red} = best, \textcolor{blue}{Blue} = second best.}
\label{tab:ddglora}
\resizebox{\columnwidth}{!}{
\begin{tabular}{l|c|c|c|c|c}
\toprule
\textbf{Training LQ Source} & \textbf{FID}$\downarrow$ & \textbf{NIQE}$\downarrow$ & \textbf{PI}$\downarrow$ & \textbf{Tenengrad}$\uparrow$ & \textbf{Entropy}$\uparrow$ \\
\midrule
realesrgan\_deg~\cite{realesrgan} & \textcolor{blue}{127.01} & 6.038 & 5.598 & 0.0608 & 6.233 \\
physics\_guided\_deg~\cite{khan2026_blind_us} & 127.17 & \textcolor{red}{\textbf{5.485}} & \textcolor{blue}{5.198} & \textcolor{blue}{0.0618} & \textcolor{blue}{6.320} \\
usbsr\_deg~\cite{usbsr} & 156.07 & 6.824 & 6.635 & 0.0405 & 6.287 \\
ultrasound\_deg & 185.75 & 7.909 & 7.377 & 0.0336 & 6.163 \\
\textbf{USenhance2023-Aligned (ours)} & \textcolor{red}{\textbf{99.00}} & \textcolor{blue}{5.619} & \textcolor{red}{\textbf{4.764}} & \textcolor{red}{\textbf{0.1043}} & \textcolor{red}{\textbf{6.861}} \\
\bottomrule
\end{tabular}
}
\end{center}
\end{table}

% To confirm that these gains are not specific to the DDG-LoRA architecture, we repeated the experiment using the CNN-based USBSR as an alternative backbone. Retraining USBSR on our dataset yielded consistent improvements in ENL, Entropy, and NIQE. This cross-architecture consistency demonstrates that the performance boost is primarily driven by the realism of the style-driven training data rather than the enhancement network itself.

%================================================================
\subsection{Comparison with State-of-the-Art Methods}

\begin{figure*}[!t]
\centering
\begin{subfigure}[t]{0.18\textwidth}
  \includegraphics[width=\linewidth]{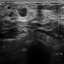}
  \caption{LQ}
\end{subfigure}
\begin{subfigure}[t]{0.18\textwidth}
  \includegraphics[width=\linewidth]{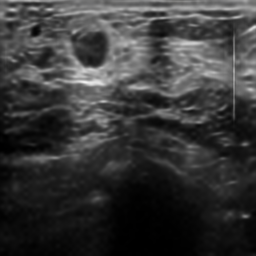}
  \caption{EDSR}
\end{subfigure}
\begin{subfigure}[t]{0.18\textwidth}
  \includegraphics[width=\linewidth]{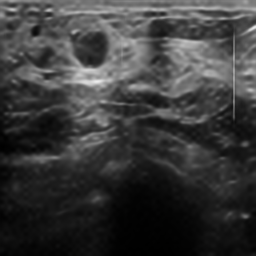}
  \caption{CycleSR}
\end{subfigure}
\begin{subfigure}[t]{0.18\textwidth}
  \includegraphics[width=\linewidth]{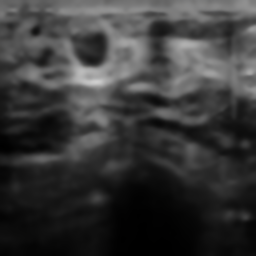}
  \caption{DARSR}
\end{subfigure}
\begin{subfigure}[t]{0.18\textwidth}
  \includegraphics[width=\linewidth]{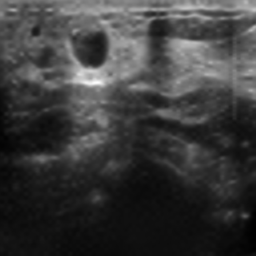}
  \caption{PDM}
\end{subfigure}\\[2pt]
\begin{subfigure}[t]{0.18\textwidth}
  \includegraphics[width=\linewidth]{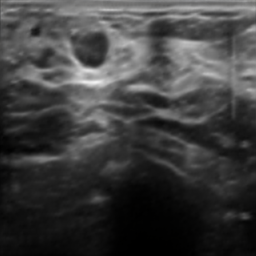}
  \caption{USBSR}
\end{subfigure}
\begin{subfigure}[t]{0.18\textwidth}
  \includegraphics[width=\linewidth]{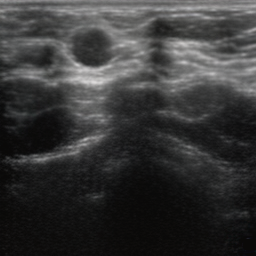}
  \caption{OSEDiff}
\end{subfigure}
\begin{subfigure}[t]{0.18\textwidth}
  \includegraphics[width=\linewidth]{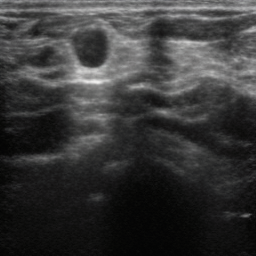}
  \caption{PiSA-SR}
\end{subfigure}
\begin{subfigure}[t]{0.18\textwidth}
  \includegraphics[width=\linewidth]{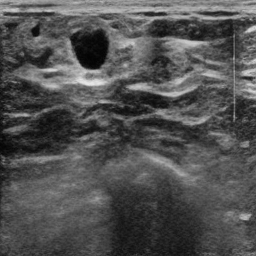}
  \caption{DDG-LoRA (ours)}
\end{subfigure}
\begin{subfigure}[t]{0.18\textwidth}
  \includegraphics[width=\linewidth]{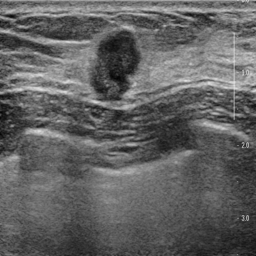}
  \caption{GT}
\end{subfigure}
\caption{Visual comparison of all methods in Table~\ref{tab:overall} on a real-world USenhance2023 sample. \textcolor{black}{The LQ input is shown in (a), the enhanced output of each method in (b)--(i), and the GT reference in (j).} \textcolor{black}{Note that because both the LQ and the GT are real-world scans, they are not pixel-aligned.}}
\label{fig:visual_overall}
\end{figure*}

We further compare our proposed DDG-LoRA against seven baselines representing different paradigms. EDSR~\cite{edsr} is a conventional CNN-based SR model. CycleSR~\cite{cyclesr} performs unsupervised SR through an indirect supervised path with cycle consistency. DARSR~\cite{darsr} tackles blind SR by regressing a degradation-adaptive correction filter, whereas PDM~\cite{pdm} explicitly learns the underlying degradation distribution. USBSR~\cite{usbsr} is specifically designed for handheld ultrasound image enhancement, employing a two-stage degradation-based unpaired training scheme. OSEDiff~\cite{osediff} and PiSA-SR~\cite{pisasr} are both one-step diffusion-based SR models built on a pretrained stable diffusion backbone, with OSEDiff focusing on efficient real-world SR and PiSA-SR introducing a dual-LoRA design for pixel-level and semantic-level adjustability.

\begin{table}[t]
\begin{center}
\caption{Overall comparison of different enhancement methods on the real-world USenhance2023 test set. \textcolor{red}{Red} = best, \textcolor{blue}{Blue} = second best.}
\label{tab:overall}
\resizebox{\columnwidth}{!}{
\begin{tabular}{l|c|c|c|c|c}
\toprule
\textbf{Method} & \textbf{FID}$\downarrow$ & \textbf{NIQE}$\downarrow$ & \textbf{PI}$\downarrow$ & \textbf{Tenengrad}$\uparrow$ & \textbf{Entropy}$\uparrow$ \\
\midrule
EDSR~\cite{edsr}           & 187.57 & 9.058 & 7.541 & 0.0573 & 6.360 \\
CycleSR~\cite{cyclesr}     & 192.36 & 8.293 & 7.139 & 0.0489 & 6.512 \\
DARSR~\cite{darsr}         & 235.03 & 8.883 & 7.466 & \textcolor{blue}{0.0680} & 6.341 \\
PDM~\cite{pdm}             & 183.92 & 8.236 & 7.416 & 0.0317 & 6.415 \\
USBSR~\cite{usbsr}         & 168.33 & 8.416 & 7.609 & 0.0426 & \textcolor{blue}{6.530} \\
OSEDiff~\cite{osediff}     & 146.80 & 6.305 & 5.837 & 0.0531 & 6.273 \\
PiSA-SR~\cite{pisasr}      & \textcolor{blue}{118.87} & \textcolor{blue}{5.763} & \textcolor{blue}{5.632} & 0.0534 & 6.252 \\
\textbf{DDG-LoRA (ours)}   & \textcolor{red}{\textbf{99.00}} & \textcolor{red}{\textbf{5.619}} & \textcolor{red}{\textbf{4.764}} & \textcolor{red}{\textbf{0.1043}} & \textcolor{red}{\textbf{6.861}} \\
\bottomrule
\end{tabular}
}
\end{center}
\end{table}

\textcolor{black}{Table~\ref{tab:overall} summarizes the quantitative comparison on the real-world USenhance2023 test set.} The non-diffusion baselines (EDSR, CycleSR, DARSR, PDM, and USBSR) consistently lag behind on most metrics, reflecting their limited capacity to model complex real-world degradations. Among them, USBSR delivers the relatively strongest numbers, since it is specifically designed for handheld ultrasound image enhancement. Within the diffusion-based group, PiSA-SR slightly outperforms OSEDiff thanks to its design. In contrast, our DDG-LoRA consistently outperforms every baseline across all five metrics, achieving the best FID (99.00), NIQE (5.619), PI (4.764), Tenengrad (0.1043), and Entropy (6.861). The FID score deserves particular attention, since FID measures the distributional distance between enhanced outputs and \textcolor{black}{authentic HQ scans}, and therefore serves as a holistic indicator of perceptual realism. Our FID of 99.00 represents a 16.7\% reduction over \textcolor{black}{the strongest baseline} (118.87), indicating that the distribution of DDG-LoRA's {enhanced outputs} is closest to that of {authentic HQ scans}, which is precisely the property our framework is designed to optimize.

{Fig.~\ref{fig:visual_overall} presents the corresponding qualitative comparison on a real-world USenhance2023 sample. EDSR, CycleSR, and PDM produce textures that diverge from the GT and exhibit obvious artifacts. DARSR is overly conservative and yields a blurry output. USBSR sharpens slightly more than DARSR but remains noticeably blurry. OSEDiff and PiSA-SR strike a better balance between fidelity and realism, yet their texture details remain visibly different from the GT. Our DDG-LoRA most closely matches the GT in characteristics such as grayscale intensity distribution, contrast, and texture detail. This qualitative observation is consistent with the quantitative comparison.}

%================================================================
\subsection{Ablation Study: Effect of Degradation Conditioning}

We perform an ablation study to isolate the contribution of the degradation-conditioning term $C(\boldsymbol{d})$ by comparing the full DDG-LoRA model against a variant in which each DG-LoRA module is reverted to a standard LoRA. Both models are trained on the USenhance2023-Aligned dataset under identical experimental settings to ensure a fair comparison.

\begin{table}[t]
\begin{center}
\caption{Ablation study on the degradation conditioning module in DDG-LoRA. \textcolor{red}{\textbf{Bold red}} indicates the better result for each metric.}
\label{tab:ablation}
\resizebox{\columnwidth}{!}{
\begin{tabular}{l|c|c|c|c|c}
\toprule
\textbf{Configuration} & \textbf{FID}$\downarrow$ & \textbf{NIQE}$\downarrow$ & \textbf{PI}$\downarrow$ & \textbf{Tenengrad}$\uparrow$ & \textbf{Entropy}$\uparrow$ \\
\midrule
w/o DDG-LoRA & \textcolor{red}{\textbf{98.98}} & \textcolor{red}{\textbf{5.589}} & 4.815 & 0.1006 & 6.808 \\
\textbf{w/ DDG-LoRA (ours)}  & 99.00 & 5.619 & \textcolor{red}{\textbf{4.764}} & \textcolor{red}{\textbf{0.1043}} & \textcolor{red}{\textbf{6.861}} \\
\bottomrule
\end{tabular}
}
\end{center}
\end{table}

The quantitative results in Table~\ref{tab:ablation} show that adding the DDG module yields better scores on PI, Tenengrad, and Entropy, while FID stays essentially unchanged at 99.00 against 98.98 and NIQE narrowly trails at 5.619 against 5.589. On balance, the gains from DDG-LoRA outweigh this small loss, confirming that the module indeed contributes to ultrasound image restoration.

%================================================================
\section{Conclusion}
In this paper, we propose a two-stage framework to train an enhancer from an unpaired LQ--HQ dataset. This method addresses the lack of pixel-aligned LQ--HQ pairs and the domain gap left by synthetic degradation in supervised handheld ultrasound enhancement. Stage one trains a CycleDiff-based style-transfer model on unaligned clinical data to synthesize pixel-aligned pairs. Stage two introduces DDG-LoRA, which extends PiSA-SR by inserting a degradation-conditioned correction matrix $C(\boldsymbol{d})$ between $A$ and $B$ in each LoRA branch to adapt the enhancement to input-specific speckle and blur levels.

A comprehensive evaluation on the real-world USenhance2023 test set demonstrates the effectiveness of both contributions. On the FID metric, \textcolor{black}{our method yields a 16.7\% reduction over the strongest baseline}. Ablation study points out that DDG-LoRA provides a further 3.7\% Tenengrad gain, and the combined system ranks first on all five no-reference metrics against seven representative baselines.

% \textbf{Limitations and future work.} The degradation descriptor is low-dimensional and may miss spatially-variant artifacts. Evaluation on downstream clinical tasks such as lesion detection remains future work, as does transfer of the learned-degrader idea to other modalities where paired data is similarly scarce.
\section*{Acknowledgment}

This work was supported by the Industrial Technology Research Institute (ITRI) and National Yang Ming Chiao Tung University Joint Research Program, under the ITRI Advanced Research Program (Project Code: Q301ARG00A).

\printbibliography

\end{document}